\documentclass{article}

\usepackage{graphicx}
\graphicspath{ {images/} }

\usepackage[final]{nips_2017_arxiv}

\usepackage[utf8]{inputenc} 
\usepackage[T1]{fontenc}    
\usepackage{hyperref}       
\usepackage{url}            
\usepackage{booktabs}       
\usepackage{amsfonts}       
\usepackage{nicefrac}       
\usepackage{microtype}      
\usepackage{subcaption}
\usepackage{enumitem}
\usepackage{multirow}
\usepackage{tabls}
\usepackage{colortbl}
\usepackage{xcolor}

\newcommand\blfootnote[1]{%
  \begingroup
  \renewcommand\thefootnote{}\footnote{#1}%
  \addtocounter{footnote}{-1}%
  \endgroup
}

\title{UPSET and ANGRI : Breaking High Performance Image Classifiers}

\author{
  Sayantan Sarkar\\ 
 University of Maryland, College Park, MD\\
 \texttt{ssarkar2@terpmail.umd.edu} \\
  \And
   Ankan Bansal \\
  University of Maryland, College Park, MD\\
  \texttt{ankan@terpmail.umd.edu} \\
  \AND
   Upal Mahbub \\
  University of Maryland, College Park, MD \\
 \texttt{umahbub@terpmail.umd.edu} \\
   \And
  Rama Chellappa\\
  University of Maryland, College Park, MD \\
   \texttt{rama@umiacs.umd.edu}\\
}

\begin{document}

\maketitle

\blfootnote{The authors are with the Department of Electrical and Computer Engineering and the Center for Automation Research, UMIACS, University of Maryland, College Park, MD 20742}

\begin{abstract}
In this paper, targeted fooling of high performance image classifiers is achieved by developing two novel attack methods. The first method generates universal perturbations for target classes and the second generates image specific perturbations. Extensive experiments are conducted on MNIST and CIFAR10 datasets to provide insights about the proposed algorithms and show their effectiveness.
\end{abstract}
\section{Introduction}
There has been a recent interest in `breaking' neural networks by generating adversarial examples which cause trained networks to misclassify. This is an important direction of research as it helps one identify vulnerabilities in the system before malicious attacks can be launched to exploit them. Adversarial samples also help in generating more varied training data, which makes deep classifiers robust \cite{goodfellow2014explaining}, \cite{kurakin2016adversarial}.

Let $\mathbf{x}$ be an image drawn from the distribution on which a classifier $C$ is trained. The semantic class of image $\mathbf{x}$ is denoted by $c_x \in \{0,1,\ldots,n-1\}$, where  $n$ denotes the total number of classes. The output of $C$ is a discrete probability distribution $\mathbf{p}\in[0,1]^n$, where the probability of class $j\in \{0,1,\ldots,n-1\}$ is $p[j]$. Let, $\mathbf{\hat{x}}$ be the corresponding transformed image, which attempts to fool $C$. It is desired that visually $\mathbf{x}$ and $\mathbf{\hat{x}}$ look similar, i.e., $\mathbf{x} \approx \mathbf{\hat{x}}$.  Thus, the definition of `targeted fooling' to a target class $t$ is given by
\begin{equation} 
argmax(C(\mathbf{\hat{x}}))=t \neq c_x, \  \mathbf{x} \approx \mathbf{\hat{x}}. 
\label{eq1} 
\end{equation}
The problem of `misclassification', defined as
\begin{equation} 
argmax(C(\mathbf{\hat{x}}))\neq c_x, \  \mathbf{x} \approx \mathbf{\hat{x}},
\label{eq2} 
\end{equation}
is a weaker problem since successful `targeted fooling' implies successful `misclassification'.

In this paper the problem specified by eq.\ref{eq1} is addressed by two proposed algorithms that fool classifiers by: 1) learning \textbf{U}niversal \textbf{P}erturbations for \textbf{S}teering to \textbf{E}xact \textbf{T}argets (UPSET); and 2) distorting images using an \textbf{A}ntagonistic \textbf{N}etwork for \textbf{G}enerating \textbf{R}ogue \textbf{I}mages (ANGRI). 
 
\section{Related Works}
Development of image perturbation algorithms for fooling deep neural networks can be very useful for generating more effective training samples and for finding flaws in trained models \cite{baluja2017adversarial}. The algorithms can be divided according to the following criteria:
\begin{enumerate}[leftmargin=5ex, nolistsep]
\item \textit{Coherency of Input Images}: One of the most popular works demonstrating the weaknesses of Deep Convolutional Neural Networks (DCNN) is \cite{nguyen2015deep}. The authors use Genetic Algorithms to generate images that have no semantic meaning to human observers, but which can make deep CNN classifiers predict classes with high confidence. However since this model generates semantically incoherent images, they are easily identifiable by humans as adversarial. The next class of algorithms transform images by adding visually imperceptible residuals such as in \cite{baluja2017adversarial}, \cite{goodfellow2014explaining}, \cite{moosavi2016universal}, \cite{Szegedy2013IntriguingPropertiesNN}, \cite{universalsegmentation}. The algorithms generate tampered images that match the input image visually, but still fool the classifier. Similar to the later algorithms, the proposed methods in this paper also generate perturbed images that are visually similar to the input images.   

\item \textit{Victim access}: In \cite{goodfellow2014explaining}, \cite{Szegedy2013IntriguingPropertiesNN}, \cite{Papernot2016TheLO}, \cite{DeepFool_moosavi2016} etc, the adversarial algorithms need knowledge of the internals of the victim model (white-box), as opposed to \cite{baluja2017adversarial}, \cite{Papernot2017PracticalBlackBoxAttack}, \cite{Kurakin2016AdversarialExamples}, \cite{moosavi2016universal}, and \cite{TransferabelAdvExp_LiuCLS16} which are black-box attack models. The proposed methods are able to mount a black-box attack.

\item \textit{Cross model generalization}: Algorithms in \cite{Papernot2016TheLO}, \cite{DeepFool_moosavi2016} target a particular classifier while the methods in \cite{nguyen2015deep}, \cite{moosavi2016universal}, \cite{xie2017adversarial}, \cite{Szegedy2013IntriguingPropertiesNN}, and \cite{TransferabelAdvExp_LiuCLS16} generalize to other classifiers which they have not been trained for. The methods described in this paper generalize across classifiers with similar structures.

\item \textit{Pattern universality}: Most algorithms generate different perturbations for different inputs, but \cite{moosavi2016universal} generates a single perturbation which universally causes incorrect classification. One of the methods in this paper produces perturbations universal to each class.

\item \textit{Targeted fooling}: Algorithms such as \cite{moosavi2016universal} consider a weak version of `fooling' where it is enough to cause the classifier to misclassify. However \cite{baluja2017adversarial}, \cite{Papernot2016TheLO}, \cite{universalsegmentation}, \cite{AdversarialExamples_KosFS17} tackle a more difficult problem, where the classifier has to predict a particular target class; the two methods proposed in this paper belong to the latter.

\item \textit{Multiple victims}: Some schemes, like \cite{baluja2017adversarial} and the methods described here, can target multiple victim classifiers together by explicitly training against them simultaneously, while others such as \cite{Papernot2017PracticalBlackBoxAttack}, \cite{moosavi2016universal}, \cite{nguyen2015deep} etc. can target only one at a time.

\end{enumerate}

\section{Proposed Algorithms}

In this work, given an input image, a target class and a pre-trained classifier, the goal is to generate an image that looks similar to the input but fools the classifier by making it predict the target class as the output, as summarized in eq. \ref{eq1}. In this section, two black-box-attack algorithms, UPSET and ANGRI are proposed for `targeted fooling'.

\subsection{UPSET: Universal Perturbations for Steering to Exact Targets}

In an $n$ class setting, UPSET seeks to produce $n$ universal perturbations $\mathbf{r}_j$, $j\in \{1, 2, \ldots, n\}$, such that 
when $\mathbf{r}_j$ is added to any image not in class $j$, the classifier will classify the resulting image as being from class $j$. The main workhorse of UPSET is a residual generating network $R$, which takes as input a target class $t$ and produces a perturbation $\mathbf{r}_t=R(t)$, which is of the same dimension as the input image $\mathbf{x}$. The adversarial image $\mathbf{\hat{x}}$ is generated by 
\begin{equation} 
\mathbf{\hat{x}} =U(\mathbf{x},t)= \max{(\min{(s \times R(t) + \mathbf{x}, 1)}, -1)}, 
\label{eqUPSET}
\end{equation}
where $U$ denotes the UPSET network. The pixel values in $\mathbf{x}$ are normalized to lie in $[-1,1]$. The network $R$ also produces values in the range $[-1,1]$. The output of $R$ is multiplied with a scaling factor $s$ as shown in eq. \ref{eqUPSET}. Setting $s=2$ ensures that $s\times R(t)$ is able to transform any value in the input space $[-1,1]$ to any value in $[-1,1]$ when added to $\mathbf{x}$. But lower values of $s$ may be chosen, which limit the maximum possible residual value at every pixel. After the addition, the output is in the range of $[-(s+1), s+1]$. It is clipped to $[-1,1]$ to produce a valid image. This adversarial image generation process is shown in fig. \ref{fig:UPSEtTraining}.

\subsection{ANGRI: Antagonistic Network for Generating Rogue Images}
ANGRI takes an input image, $\mathbf{x}$, belonging to class $c_x$, and a target class, $t\neq c_x$, and transforms it into a new image, $\mathbf{\hat{x}}$, such that the classifier mislabels it as being an object from class $t$. Compared to UPSET, ANGRI does not produce universal perturbations, as its output depends on the input image. The transformed image is 
$\mathbf{\hat{x}} =A(\mathbf{x},t)$, where $A$ denotes the ANGRI network.

\subsection{Loss Function}
\begin{figure}
    \centering
    \begin{subfigure}[b]{0.49\textwidth}
        \includegraphics[width=\textwidth]{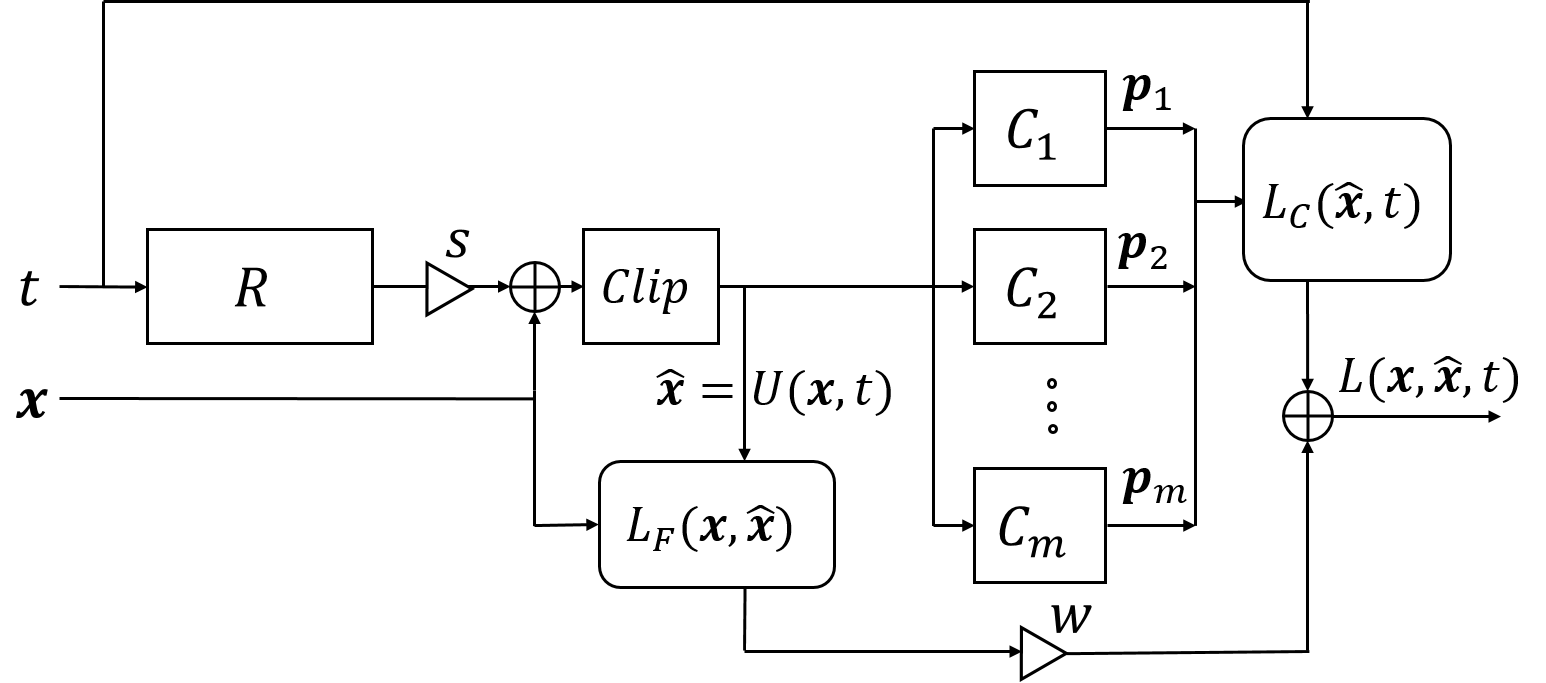}
        \caption{Training scheme for UPSET($U$).}
        \label{fig:UPSEtTraining}
    \end{subfigure}
    \begin{subfigure}[b]{0.49\textwidth}
        \includegraphics[width=\textwidth]{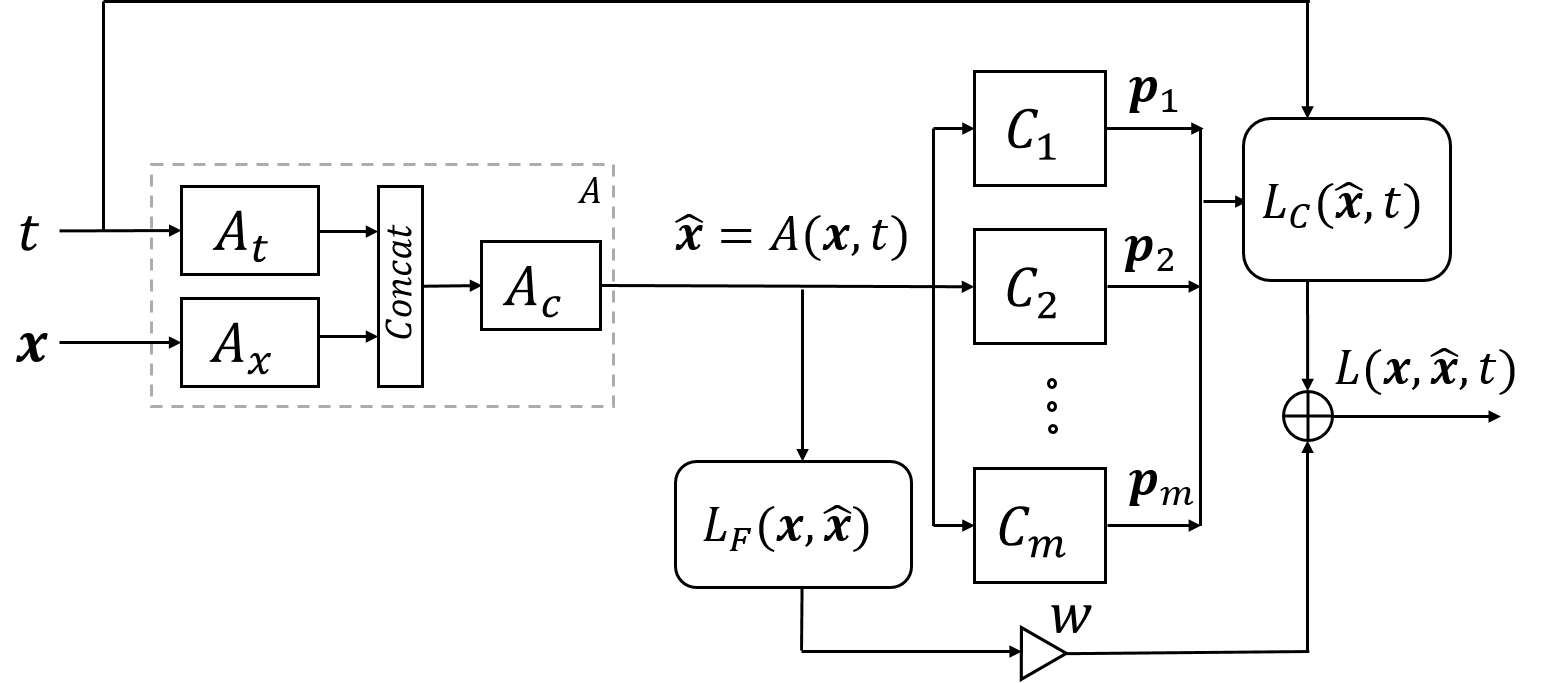}
        \caption{Training scheme for ANGRI($A$).}
        \label{fig:ANGRITraining}
    \end{subfigure}
\caption{Training networks for generating adversarial images. $\mathbf{x}$ is the input image, $t$ is the target class, and $\mathbf{\hat{x}}$ is the output adversarial image.}
\label{fig:TwoTrainingSchemes}
\end{figure}

Given $m$ pretrained classifiers $C_i$, denote their output classification probabilities for an adversarial image $\mathbf{\hat{x}}$ by $\mathbf{p}_i$, that is, $\mathbf{p}_i = C_i(\mathbf{\hat{x}})$. Then loss, for both UPSET and ANGRI, is defined in eq. \ref{loss}.
\begin{eqnarray} 
L(\mathbf{x},\mathbf{\hat{x}},t) &=& L_C(\mathbf{\hat{x}}, t)+L_F(\mathbf{x}, \mathbf{\hat{x}}) = -\sum_{i=1}^{m} {log(C_i(\mathbf{\hat{x}})[t])} \ + \ w\parallel \mathbf{\hat{x}} - \mathbf{x}\parallel _k^k,
\label{loss}
\end{eqnarray}
where, $L_C$ denotes (mis)classification loss and $L_F$ denotes fidelity loss. $L_C$ is a categorical cross-entropy loss that penalizes the generator network if the classifier does not predict the target class, $t$, and $L_F$ is the norm of the differences between $\mathbf{x}$ and $\mathbf{\hat{x}}$ which ensures that the input and output images look similar. Weight $w$ is used to trade off between fidelity and fooling capability of the generated adversarial example. The choice of $k$ should be such that it does not promote sparsity, else the residuals will accumulate in small regions and will be conspicuous. It suffices to use $k=2$, that is, the L2 norm. In the case of UPSET, since the network already generates a residual, the second term in eq. \ref{loss} is substituted by $\parallel R(\mathbf{x},t)\parallel _2^2$. The training schemes of the two systems with the associated losses are shown in fig. \ref{fig:TwoTrainingSchemes}.

\section{Experimental Results}
The following four metrics are used for evaluating the performances of the proposed fooling networks:
\begin{enumerate}[leftmargin=5ex, nolistsep]
\item \textit{Targeted fooling rate} (TFR): Rate of occurrence of the event defined by eq. \ref{eq1} (Higher means that the classifier is easily fooled).
\item \textit{Misclassification rate} (MR): Rate of occurrence of the event defined by eq. \ref{eq2} (Higher means that the classifier is easily fooled).
\item \textit{Fidelity score} (FS): Mean residual norm per pixel per channel (Lower means that the residual is less conspicuous).
\item \textit{Confidence} (C): Mean probabilities of the target class when successfully fooled. This shows the confidence of the classifier when it is fooled. (Higher means it is more confident for the incorrect class).
\end{enumerate}

Training UPSET or ANGRI with multiple classifiers is referred to as `simultaneous' training. If UPSET or ANGRI is trained and tested on the same set of classifiers, it is called `self' attack, else it is called `cross' attack. The results are described on two datasets MNIST \cite{MNIST_Lecun_1998} and CIFAR-10 \cite{Krizhevsky_Cifar10Citation}.

\subsection{Network Architectures} 

This section describes the network architectures of the classifier, UPSET, and ANGRI models for both MNIST and CIFAR-10 datasets. The networks are defined using the following notation:
\begin{enumerate}[leftmargin=5ex, nolistsep]
\item \textit{Activation functions}: ReLU ($R$) or Leaky ReLU ($L$) is used everywhere except in the last layers which use softmax ($S$) or tanh ($T$) activation functions.
\item \textit{Regulatization}: $D_{\phi}$ denotes dropout with probability $\phi$ and $B$ denotes batch-normalization.
\item \textit{Identity}: No regularization or activation is denoted by the identity layer, $I$. 
\item \textit{Convolutional layer}: $C_{\mu, \nu, \zeta, \beta, \gamma}$ denotes convolutional layers with $\nu$ filters of kernel size $\mu\times \mu$ with strides $\zeta$ and followed by regularization $\beta$ and activation function $\gamma$.
\item \textit{Deconvolution layer}: $DC_{\mu, \nu, \zeta, \gamma}$ denotes a deconvolution layer with $\nu$ filters of kernel size $\mu\times \mu$ with strides $\zeta$ and followed by activation function $\gamma$
\item \textit{Dense}: $F_{\lambda, \gamma}$ denotes a fully connected layer with $\lambda$ nodes followed by activation function $\gamma$.
\item \textit{Pooling layers}: $M_{\ell, \zeta}$ denotes a max-pooling layer of kernel size $\ell\times \ell$ and stride $\zeta$, and $A$ denotes an average pooling layer.
\item \textit{Blocks}: $\textbf{R}_{\gamma}$ is a residual block similar to the one defined in \cite{he2016deep} which contains regularization $\gamma$.
\end{enumerate}

\subsubsection{MNIST} \label{mnistnetworkarch}
For MNIST experiments, three classifier networks, $M_1$, $M_2$, and $M_3$, are trained. $M_1$ is a network with $2$ convolutional layers and $2$ dense layers. $M_2$ is similar to $M_1$, but with more parameters. $M_3$ is a dense network with $3$ layers. The structures of the classifier networks are:
\begin{enumerate}[leftmargin=5ex, nolistsep]
\item $M_1: C_{3,32,1,I,R} \rightarrow C_{3,64,1,I,R} \rightarrow M_{2,1}\rightarrow F_{128,R} \rightarrow D_{0.5}\rightarrow F_{10, S}$
\item $M_2: C_{3,64,1,I,R} \rightarrow C_{3,128,1,I,R}\rightarrow M_{2, 1}\rightarrow F_{256,R} \rightarrow D_{0.5}\rightarrow F_{10, S}$
\item $M_3: F_{512,R} \rightarrow D_{0.5} \rightarrow F_{256,R} \rightarrow D_{0.5}\rightarrow F_{10, S}$
\end{enumerate}
An additional model $M_4$ with exactly the same architecture as $M_1$, is trained with additive noise in the input MNIST digit images as well as with rotation ($\pm 10$ degree), and shift ($20\%$ along both axes). The initial test set accuracy of the four models are $99.11$\%, $99.23$\% , $97.39$\% and $98.03$\% respectively. All classifiers are trained for $12$ epochs and all UPSET and ANGRI models described in the following sections are trained for $25$ epochs.

The UPSET network's residual generator $R$ (fig. \ref{fig:UPSEtTraining}) consists of six dense layers as shown below:
\begin{center}
$R: F_{128, R} \rightarrow F_{256, L} \rightarrow F_{512, L} \rightarrow F_{1024, L} \rightarrow F_{512, L} \rightarrow F_{784, T}$.
\end{center}
For ANGRI, the base network for $10$-D target, $A_t$, the base network for the $784$-D input image, $A_x$, and the top network $A_c$ (as shown in fig. \ref{fig:ANGRITraining}) are defined as:
\begin{center}
$A_t: F_{128, L} \rightarrow F_{256, L} \rightarrow F_{512, L}$, $A_x: F_{128, L} \rightarrow F_{256, L} \rightarrow F_{512, L}$,\\
$A_c: [A_t, A_I] \rightarrow F_{1024, L} \rightarrow F_{512, L} \rightarrow F_{784, T}$.
\end{center}

\subsubsection{CIFAR-10} \label{cifar10networkarch} 
Four classifiers, $C_1-C_4$ are trained. $C_1$ and $C_2$ are resnet style networks, while $C_3$ and $C_4$ are deep convolutional networks with $9$ and $4$ convolutional layers respectively. The initial test set accuracies of the four classifiers, $C_1-C_4$, are $86.65\%, 86.58\%, 86.86\%$ and $87.09\%$ respectively. The networks are shown below:
\begin{enumerate}[leftmargin=5ex, nolistsep]
\item $C_1: C_{7,64,2,B,R} \rightarrow M_{3,2} \rightarrow C_{3,64,1,I,I} \rightarrow \textbf{R}_B
    \rightarrow \textbf{R}_B \rightarrow \textbf{R}_B \rightarrow \textbf{R}_B \rightarrow \textbf{R}_B \rightarrow \textbf{R}_B \rightarrow \textbf{R}_B \rightarrow B \rightarrow R \rightarrow A
    \rightarrow F_{10, S}$ 
\item $C_2: C_{7,64,2,D_{0.5},R} \rightarrow M_{3,2} \rightarrow C_{3,64,1,I,I} \rightarrow \textbf{R}_{D_{0.3}}
   \rightarrow \textbf{R}_{D_{0.3}} \rightarrow \textbf{R}_{D_{0.3}} \rightarrow \textbf{R}_{D_{0.3}} \rightarrow \textbf{R}_{D_{0.3}} \rightarrow \textbf{R}_{D_{0.3}} \rightarrow \textbf{R}_{D_{0.3}} \rightarrow B \rightarrow R \rightarrow A \rightarrow F_{10,S}$
\item $C_3: C_{3,32,1,I,L} \rightarrow C_{3,32,1,I,L} \rightarrow M_{2,2} \rightarrow C_{3,64,1,I,L} \rightarrow C_{3,128,1,I,L} \rightarrow  D_{0.5} \rightarrow C_{3,256,1,I,L} \rightarrow M_{2,2} \rightarrow D_{0.5} \rightarrow
C_{3,128,1,I,L} \rightarrow D_{0.5} \rightarrow C_{3,64,1,I,L} \rightarrow M_{2,2} \rightarrow C_{3,32,1,I,L} \rightarrow D_{0.5} \rightarrow C_{3,16,1,I,L} \rightarrow D_{0.5} \rightarrow F_{256,L} \rightarrow D_{0.5}
    \rightarrow F_{10,S}$
\item $C_4: C_{3,32,1,I,R} \rightarrow C_{3,32,1,I,R} \rightarrow M_{2,2}
    \rightarrow D_{0.5} \rightarrow C_{3,64,1,I,R} \rightarrow C_{3,64,1,I,R}
    \rightarrow M_{2,2} \rightarrow D_{0.5} \rightarrow F_{512,R} \rightarrow
    D_{0.5} \rightarrow F_{10,S}$    

   \end{enumerate}

The UPSET network $R$ (fig. \ref{fig:UPSEtTraining}) for CIFAR-10 is a network with $6$ deconvolutional layers, followed by $6$ convolutional layers, as defined below:
\begin{center}
$R: DC_{3, 32, 2, L} \rightarrow DC_{3, 64, 2, L} \rightarrow DC_{3, 64, 2, L} \rightarrow DC_{3, 128, 2, L} \rightarrow DC_{3, 128, 2, L} \rightarrow DC_{3, 128, 2, L} \rightarrow C_{3, 128, 1, I, L} \rightarrow C_{3, 64, 2, 1, L} \rightarrow C_{1, 32, 1, I, L} \rightarrow C_{1, 16, 1, I, L} \rightarrow C_{1, 8, 1, I, L} \rightarrow C_{1, 3, 1, I, T}$.
\end{center}

The ANGRI network for CIFAR-10 has $3$ sections as shown in fig. \ref{fig:ANGRITraining}. 
Both the $A_t$ network, which takes a $10$-D one-hot target vector as input, and the $A_x$ network, which takes a $3\times 32 \times 32$ image as input, have $3$ deconvolutional layers. The $A_c$ network after the merging of $A_t$ and $A_x$ has $3$ deconvolutions, followed by 6 convolutional layers. The exact structure is summarized below:
\begin{center}
$A_t: DC_{3, 32, 2, L} \rightarrow DC_{3, 64, 2, L} \rightarrow DC_{3, 64, 2, L}$, 
$A_x: DC_{3, 32, 1, L} \rightarrow DC_{3, 64, 2, L} \rightarrow DC_{3, 128, 2, L}$,\\
$A_c: [A_t, A_I] \rightarrow DC_{3, 128, 2, L} \rightarrow DC_{3, 128, 2, L} \rightarrow DC_{3, 128, 2, L} \rightarrow C_{3, 128, 1, I, L} \rightarrow C_{3, 64, 2, I, L} \rightarrow C_{1, 32, 1, I, L} \rightarrow C_{1, 16, 1, I, L} \rightarrow C_{1, 8, 1, I, L} \rightarrow C_{1, 3, 1, I, T}$.
\end{center}

\subsection{Sample Images and Residuals}
\begin{figure}[t]
\centering
\includegraphics[width=0.8\textwidth]{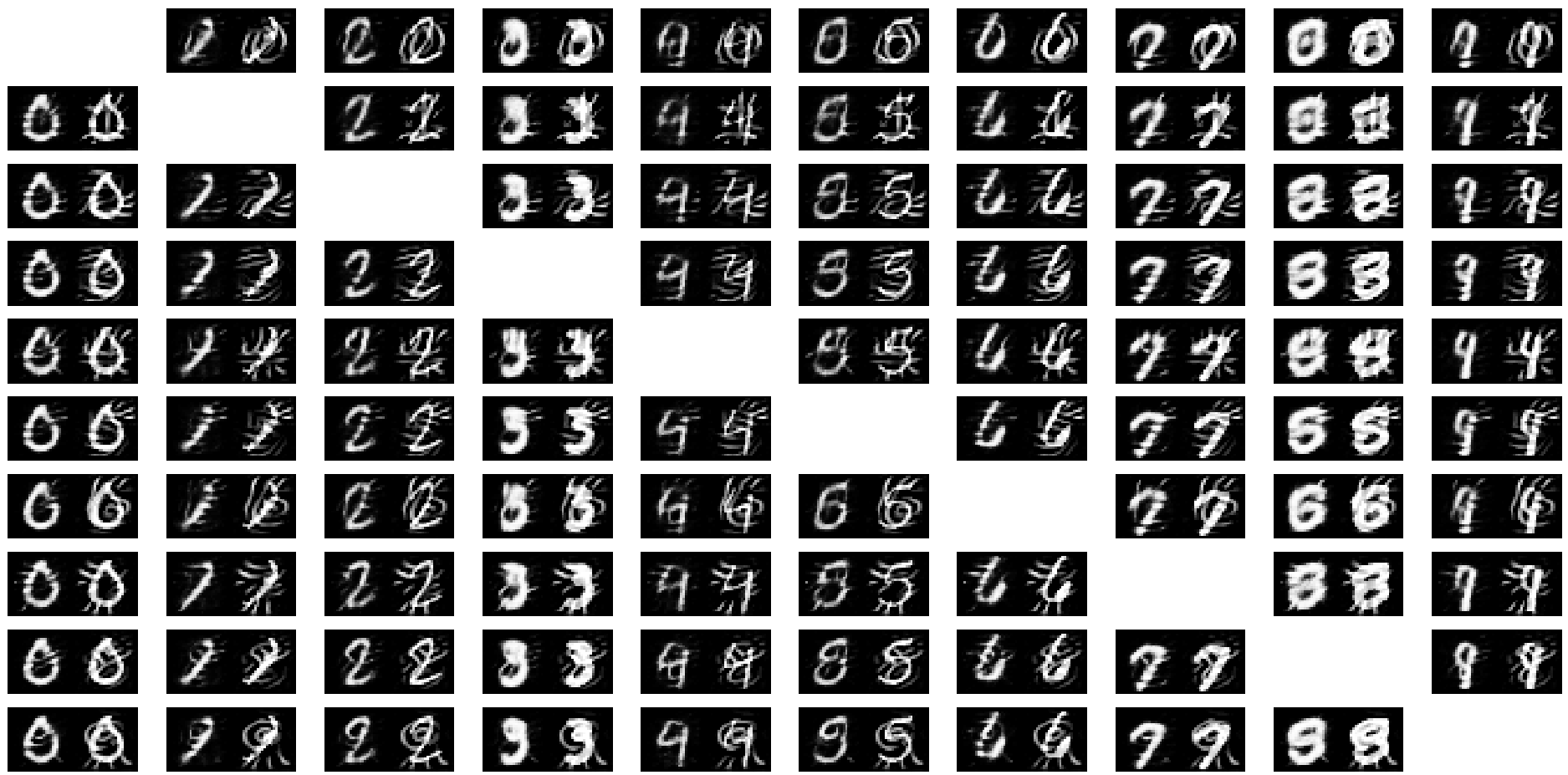}
\vskip 0pt
\caption{Sample of adversarial images from MNIST. Each column represents input classes $0$ to $9$, while each row represents target classes $0$ to $9$. The left image in each cell is generated using ANGRI ($w=0.1$), while the right is from UPSET ($w=0.06$).}
\label{mnistsamples}
\vskip -10pt
\end{figure}

\begin{figure}
    \centering
    \begin{subfigure}[b]{0.38\textwidth}
        \includegraphics[width=\textwidth]{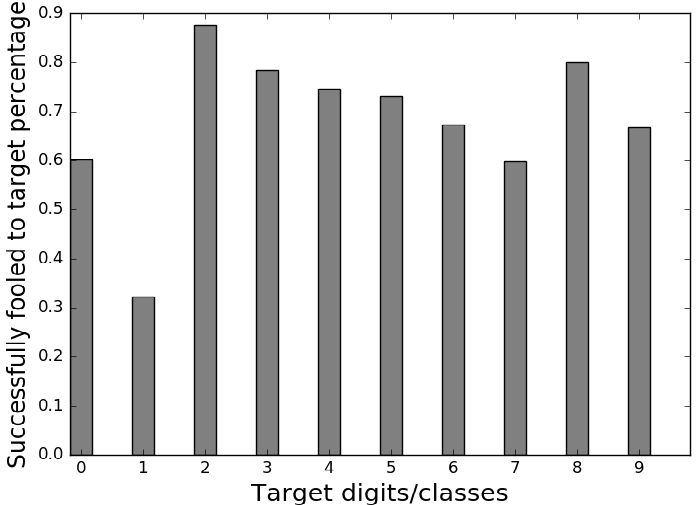}
        \caption{Fooling percentage for each target class. It is difficult to make the classifier say that an image is $0$ or $1$ when it is not so.}
        \label{perclassfool}
    \end{subfigure}
    ~ 
    \begin{subfigure}[b]{0.48\textwidth}
        \includegraphics[width=\textwidth]{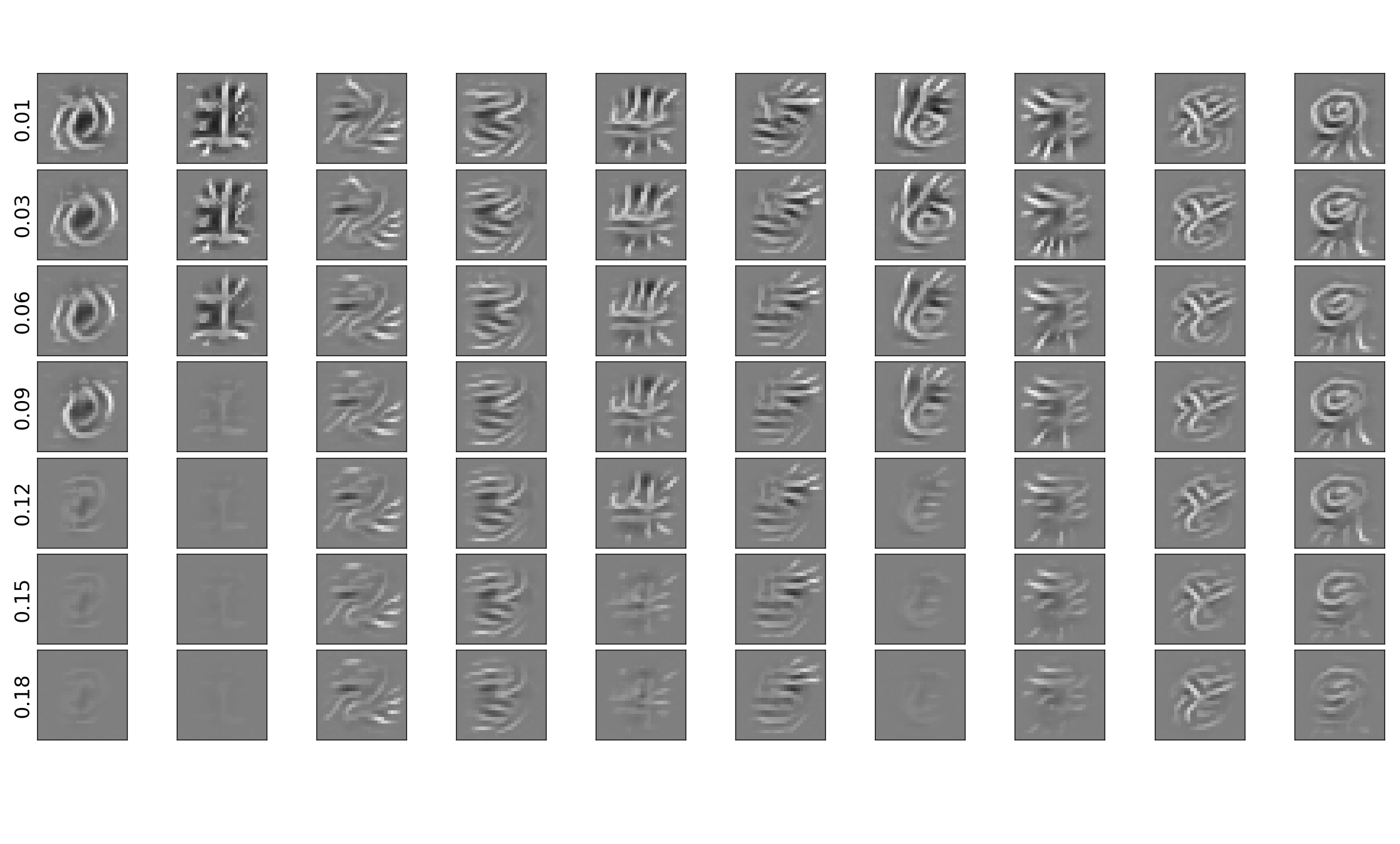}
        \caption{UPSET residuals with varying weights on visual loss for MNIST. Weights increase with rows. Columns represent each target class.}
        \label{fig:MNIST_UPSET_res}
    \end{subfigure}
\caption{Analysis of UPSET ($w=0.06,s=2$) for MNIST.}
\label{fig:upsetresis}
\end{figure}

Some sample successful adversarial images for MNIST are shown in fig. \ref{mnistsamples}, generated using $w=0.1$ for ANGRI and $w=0.06,s=2$ for UPSET. Fig. \ref{perclassfool} shows the rate of successful attacks per target class. Clearly, targets $0$ and $1$ are difficult to generate residuals for. Reflecting that fact, in fig. \ref{fig:MNIST_UPSET_res} it can be seen that as visual loss weight $w$ increases, UPSET stops generating residuals for classes $0$ and $1$. As $w$ increases UPSET is forced to add residuals of less magnitude. Therefore, it gives up on these difficult targets and reserves its visual loss budget for the classes it can successfully fool.

\begin{figure}[t]
\centering
\includegraphics[width=0.9\textwidth]{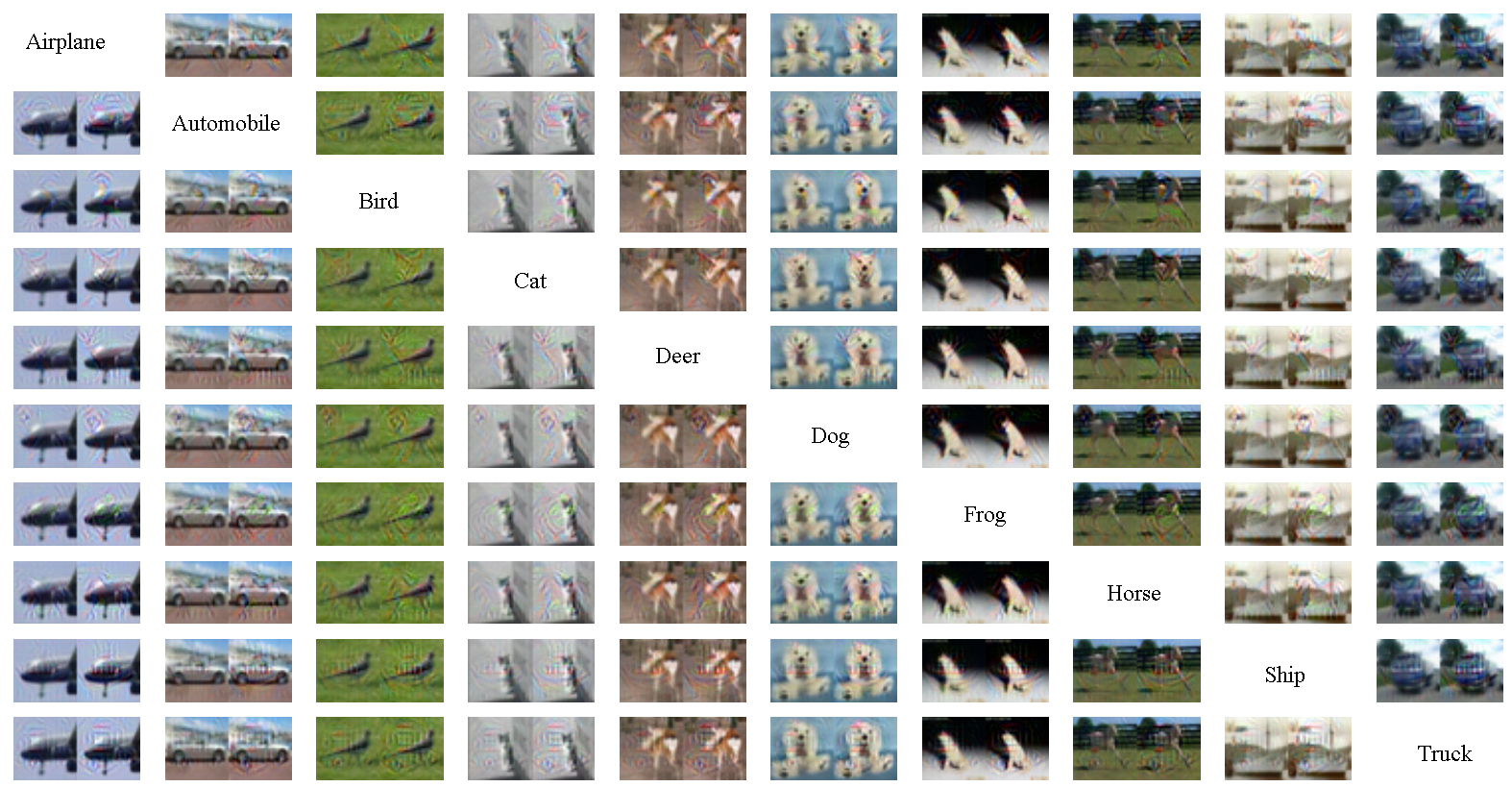}
\vskip 0pt
\caption{Sample of adversarial images from CIFAR-10. Each column represents input classes $0$ to $9$, while each row represents target classes $0$ to $9$. The left image in each cell is generated using ANGRI ($w=0.11$), while the right is from UPSET ($w=0.08$). Thus the image in row 4, column 7 looks like a frog, but is classified as a cat.(Best viewed electronically)}
\label{cifar10samples}
\vskip -10pt
\end{figure}

\begin{figure}
    \centering
    \begin{subfigure}[b]{1\textwidth}
        \includegraphics[width=\textwidth]{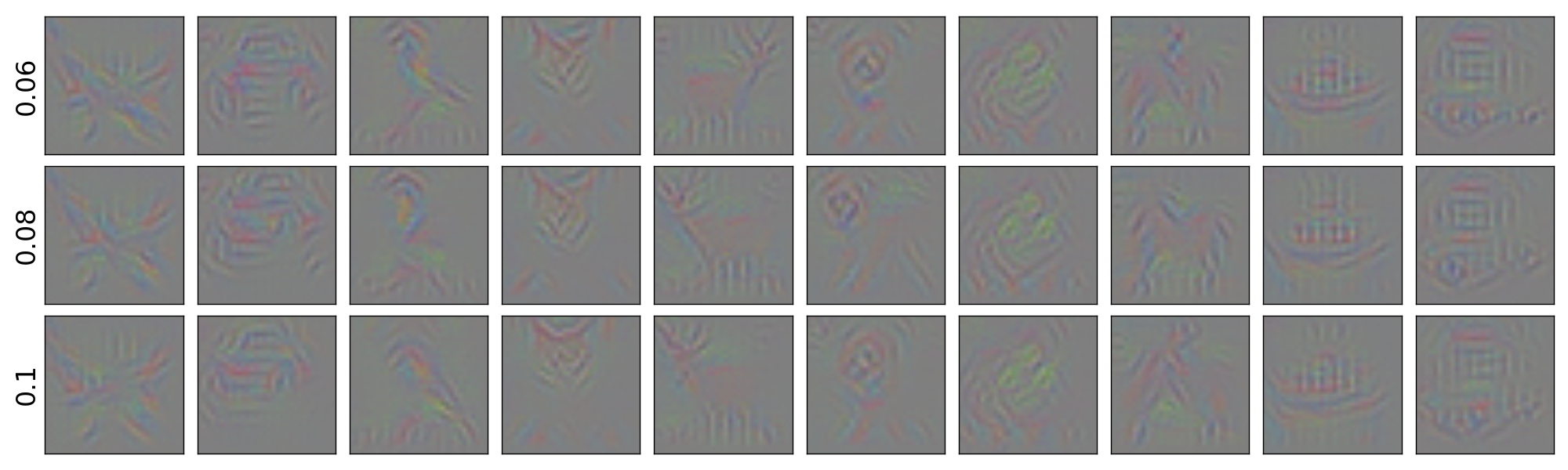}
        \caption{UPSET (trained with $C_1$) residuals with varying weights (rows) on visual loss of ten different targets (columns) for CIFAR-10.}
        \label{fig:UPSET1_res}
    \end{subfigure}
    ~ 
    \begin{subfigure}[b]{1\textwidth}
        \includegraphics[width=\textwidth]{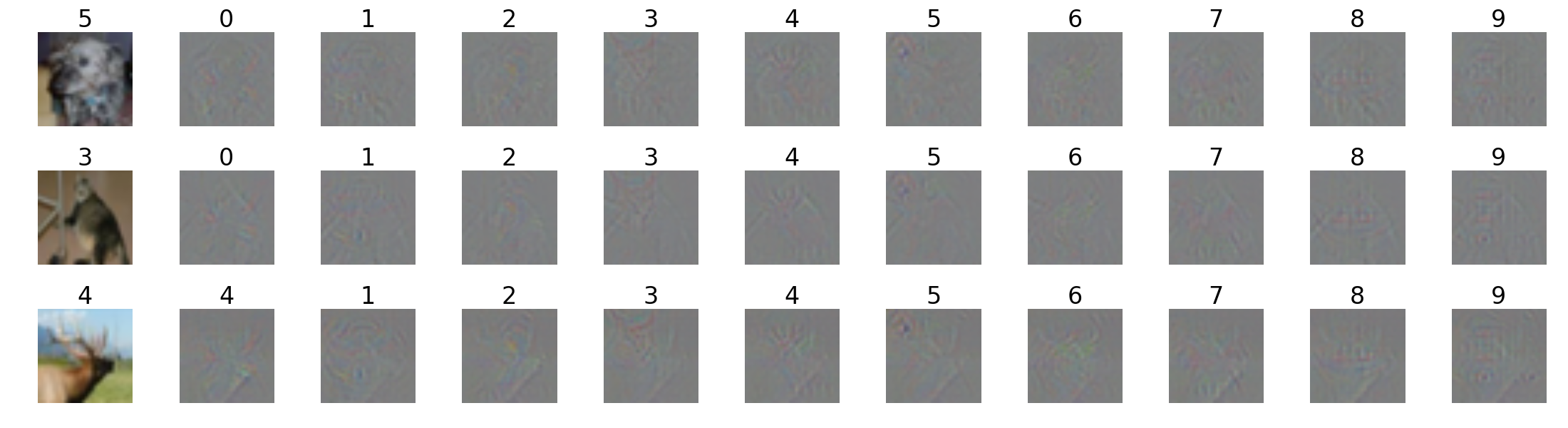}
        \caption{ANGRI (trained with $C_1$, $w=0.11$) residuals for ten different target classes (columns) and three different input images (rows) from CIFAR-10.}
        \label{fig:angri_res}
    \end{subfigure}
\caption{Residuals from UPSET for CIFAR-10. White depicts large positive values, black depicts large negative values, and gray depicts zero residual. (Best viewed electronically)}
\label{fig:cifarresis}
\end{figure}

Fig. \ref{cifar10samples} shows some sample outputs of UPSET and ANGRI for each input and target class in CIFAR-10. The difference between magnitudes of perturbations for ANGRI and UPSET can be seen clearly in fig. \ref{fig:cifarresis}. Also, in fig. \ref{fig:angri_res}, the residuals have more variations, since ANGRI produces image specific distortions, unlike UPSET which produces target specific perturbations independent of the input image. Therefore, ANGRI is able to generate more visually inconspicuous distortions as compared to UPSET as can be seen in fig. \ref{cifar10samples}.

\subsection{Effect of Visual Loss Weight}

\begin{figure}
    \centering
    \begin{subfigure}[b]{0.49\textwidth}
        \includegraphics[width=\textwidth]{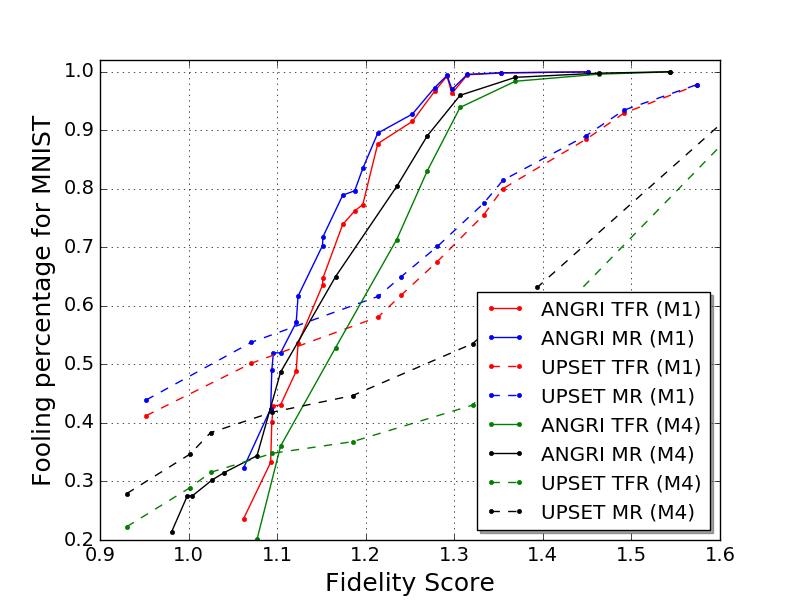}
        \caption{Fooling rates vs. fidelity scores for MNIST.}
        \label{mnistfidrate}
    \end{subfigure}
    ~ 
    \begin{subfigure}[b]{0.49\textwidth}
        \includegraphics[width=\textwidth]{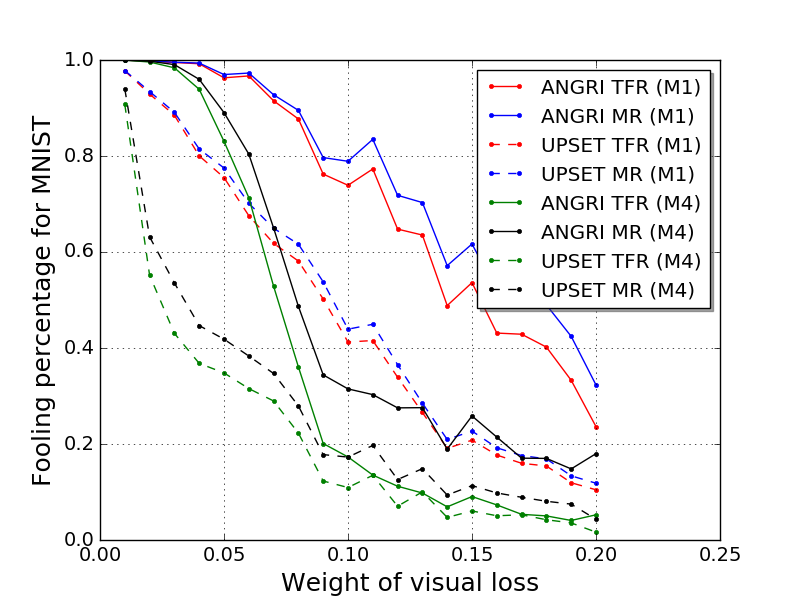}
        \caption{Fooling rates vs. Visual loss weight $w$ for MNIST.}
        \label{mnistwtrate}
    \end{subfigure}
\caption{Training networks for generating adversarial images. Increasing $w$ decreases Targeted Fooling Rate (TF) and Misclassification Rate (MR). TFR is always less than MR. ANGRI gives higher TFR for same $w$ than UPSET.}
\label{fig:FoolingRateVsVisLoss&FidScore}
\end{figure}

The effect of varying the visual loss weight $w$ is studied on MNIST. It is varied from $0.01$ to $0.2$ in steps of $0.01$. Two cases are considered, one with classifier $M_1$ and the other with $M_4$. Note that both classifiers have the same architecture, but $M_4$ is trained with augmented data as described in Section \ref{mnistnetworkarch}. Fig. \ref{mnistfidrate} shows that, as expected, the fooling rates decrease as the fidelity score increases, that is, by adding residuals with higher magnitudes to the images UPSET and ANGRI are able to fool better. Fig. \ref{mnistwtrate} shows that with increasing weight of visual loss, fooling rates go down as the network is now forced to add smaller residuals to corrupt the input image. In both images in fig. \ref{fig:FoolingRateVsVisLoss&FidScore} MR is greater than TFR, since targeted fooling is a stronger condition than misclassification. For the region of high TFR (greater than about $60\%$), ANGRI produces residuals of lower magnitude than UPSET. Also, it is more difficult to fool $M_4$ than $M_1$, which means that having noise in training data helps make the classifier more robust against UPSET or ANGRI. However, as depicted in fig. \ref{mnistfidrate}, ANGRI shows only a slight drop in performance for $M_4$, unlike UPSET.

\subsubsection{Simultaneous Training and Cross Model Generalization}
For this experiment, UPSET models are trained with $s=2$ and $w=0.06$, while ANGRI models are trained with $w=0.1$. The generalizability of the proposed methods is studied by considering their TFR when attacking a different classifier than the one it was trained on. Since network architecture might have a strong effect on generalizability, different categories of classifiers are trained. In MNIST there are two convolutional networks, $M_1$ and $M_2$ and one dense network $M_3$. In CIFAR-10 there are two resnet style networks, $C_1$ and $C_2$, and two convolutional networks, $C_3$ and $C_4$. 

Table \ref{resultstablemnist} summarizes the results for self (train and test on $C_i$) and cross (train on $C_i$ and test on $C_j$, $i \neq j$) attacks for MNIST. For CIFAR-10, table \ref{resulttablecifaronetoone} reports the self and cross attack TFRs for UPSET and ANGRI models trained with a single classifier, while table \ref{resulttablecifarpairtoone} shows the performance for simultaneous training with two classifiers. Some key observations from the tables are:

\begin{enumerate}[leftmargin=5ex, nolistsep]
\item Cross attack fooling rates are lower than self attack rates.
\item UPSET and ANGRI models trained with a classifier of a certain structure (convolutional/resnet/dense) are able to fool other classifiers of similar structures much better than those of different structures. UPSET and ANGRI models trained with resnet style classifiers seem to generalize well to other structures in CIFAR-10. However, when trained on classifiers with non-resnet structures, neither method is able to fool other types of classifiers.
\item Training UPSET or ANGRI with multiple classifiers together helps them generalize better. This is clearly seen when comparing the TFRs in tables \ref{resulttablecifaronetoone} and \ref{resulttablecifarpairtoone}. Let $C_i,C_j \rightarrow C_k$ mean that UPSET and ANGRI are trained with $C_i$ and $C_j$, and then used to attack $C_k$. In almost all cases, $C_i,C_j \rightarrow C_k$ gives higher fooling rates than either $C_i \rightarrow C_k$ or $C_j \rightarrow C_k$, when $i \neq j \neq k$. This generalization gained from simultaneous training, comes at a small cost as the performance of self attack drops a little compared to when trained separately.


\item As can be seen in table \ref{resultstablemnist}, the confidence, $C$, of the classifier when it is fooled is high.
\end{enumerate}

\begin{table}[h!]
\centering
\caption{Comparative performances of ANGRI and UPSET for different classifiers for MNIST.}
{\tablinesep=0ex
\begin{tabular}{c c c c c c c | c c c c}
\hline
\multirow{2}{*}{No.} & \multirow{2}{*}{Trained on} & \multirow{2}{*}{Victim}  & \multicolumn{4}{c}{A} & \multicolumn{4}{c}{U}\\
\cline{4-11}
& & & TFR & MR & C & FS & TFR & MR & C & FS \\
\hline
\hline
$1$ & $M_1$   & $M_1$ &  $74.70$ & $79.85$ & $0.78$ & $1.17$ & $70.53$ & $73.01$ & $0.94$ & $1.29$\\
\hline
$2$ &$M_2$   & $M_2$ &   $78.15$ & $80.38$ & $0.81$ & $1.18$ & $73.03$ & $74.08$ & $0.94$ & $1.28$ \\
\hline
$3$ &$M_3$   & $M_3$ &  $53.89$ & $60.98$ & $0.70$ & $1.10$ & $56.29$ & $59.70$ & $0.88$ & $1.09$ \\
\hline

$4$ &$M_1$   & $M_2$ &   $50.82$ & $57.32$ & $0.78$ & $1.17$ & $67.47$ & $68.73$ & $0.93$ & $1.29$ \\
\hline
$5$ &$M_2$   & $M_1$ &   $57.05$ & $65.19$ & $0.77$ & $1.18$ & $59.21$ & $63.67$ & $0.93$ & $1.28$ \\
\hline
$6$ &$M_2$   & $M_3$ &   $3.72$ & $8.77$ & $0.68$ & $1.18$ & $7.47$ & $12.30$ & $0.70$ & $1.28$ \\
\hline
$7$ &$M_3$   & $M_2$ &   $1.83$ & $4.27$ & $0.73$ & $1.10$ & $1.25$ & $2.72$ & $0.77$ & $1.08$ \\
\hline
$8$ &$M_3$   & $M_1$ &   $2.15$ & $4.91$ & $0.71$ & $1.10$ & $1.49$ & $3.40$ & $0.76$ & $1.08$\\
\hline
$9$ &$M_1$   & $M_3$ &   $3.10$ & $7.88$ & $0.68$ & $1.17$ & $9.56$ & $14.51$ & $0.72$ & $1.29$ \\
\hline

$10$ &$M_1, M_3$   & $M_1$ &   $58.21$ & $66.82$ & $0.79$ & $1.25$ & $55.38$ & $59.28$ & $0.93$ & $1.27$\\
\hline
$11$ &$M_1, M_3$   & $M_3$ &   $46.25$ & $56.91$ & $0.72$ & $1.25$ & $49.27$ & $53.40$ & $0.85$ & $1.27$ \\
\hline
$12$ &$M_2, M_1$   & $M_2$ &  $80.17$ & $84.12$ & $0.84$ & $1.24$ & $66.16$ & $69.08$ & $0.94$ & $1.28$ \\
\hline
$13$ &$M_2, M_1$   & $M_1$ & $81.22$ & $83.90$ & $0.85$ & $1.24$ & $70.84$ & $72.16$ & $0.93$ & $1.28$ \\
\hline
$14$ &$M_3, M_2$   & $M_3$ &   $54.81$ & $60.91$ & $0.76$ & $1.25$ & $55.94$ & $57.74$ & $0.92$ & $1.25$\\
\hline
$15$ &$M_3, M_2$   & $M_2$ &   $46.03$ & $54.21$ & $0.70$ & $1.25$ & $47.03$ & $51.27$ & $0.84$ & $1.25$\\
\hline

$16$ &$M_1, M_2, M_3$   & $M_1$ &   $67.23$ & $74.58$ & $0.80$ & $1.23$ & $61.24$ & $64.83$ & $0.93$ & $1.28$\\
\hline

$17$ &$M_1, M_2, M_3$   & $M_2$ &   $70.94$ & $75.84$ & $0.82$ & $1.23$ & $65.70$ & $66.88$ & $0.93$ & $1.28$\\
\hline

$18$ &$M_1, M_2, M_3$   & $M_3$ &   $33.25$ & $43.49$ & $0.69$ & $1.23$ & $37.70$ & $42.68$ & $0.81$ & $1.28$ \\
\hline
\end{tabular}
}
\vskip -5pt
\label{resultstablemnist}
\end{table}


\begin{table}[h!]
\centering
\caption{Comparative performances of ANGRI and UPSET on CIFAR-10 in terms of TFR when training on a single classifier. Each cell represents the (first element) TFR with UPSET and (second element) TFR with ANGRI. $C_1$ and $C_2$ are both resnet style while $C_3$ and $C_4$ are convolutional networks. As the bold numbers show, the fooling networks generalize better if they target networks of similar structure.}
{\tablinesep=0ex
\begin{tabular}{ c  c  c  c  c  } 
\hline
\multirow{2}{*}{Victim} & \multicolumn{4}{c}{Trained on} \\
\cline{2-5}
& $C_1$ & $C_2$ & $C_3$ & $C_4$ \\
\hline
\hline
$C_1$ & $\mathbf{91\%,94\%}$ & $\mathbf{48\%,50\%}$ & $5\%,4\%$ & $6\%,6\%$ \\
\hline
$C_2$ & $\mathbf{50\%,47\%}$ & $\mathbf{90\%,95\%}$ & $8\%,5\%$ & $11\%,14\%$ \\
\hline
$C_3$ & $46\%,44\%$ & $64\%,75\%$ & $\mathbf{97\%,98\%}$ & $\mathbf{67\%,75\%}$ \\
\hline
$C_4$ & $40\%,36\%$ & $61\%,71\%$ & $\mathbf{39\%,31\%}$ & $\mathbf{93\%,98\%}$ \\
\hline
\end{tabular}
}
\vskip -5pt
\label{resulttablecifaronetoone}
\end{table}

\begin{table}[h!]
\centering
\caption{Comparative performances of ANGRI and UPSET on CIFAR-10 in terms of TFR when training on 2 classifiers. Each cell
represents the (first element) fooling rate with UPSET and (second element) fooling rate with ANGRI.}
{\tablinesep=0ex
\begin{tabular}{ c  c  c  c  c  c  c }
\hline
\multirow{2}{*}{Victim} & \multicolumn{6}{c}{Trained on} \\
\cline{2-7}
& $C_1,C_2$ & $C_1,C_3$ & $C_1,C_4$ & $C_2,C_3$ & $C_2,C_4$ & $C_3,C_4$ \\
\hline
\hline
$C_1$ & $87\%,96\%$ & $87\%,91\%$ & $85\%,92\%$ & $51\%,44\%$ & $45\%,48\%$ & $10\%,8\%$\\
\hline
$C_2$ & $85\%,96\%$ & $54\%,60\%$ & $59\%,68\%$ & $88\%,92\%$ & $85\%,93\%$ & $18\%,14\%$\\
\hline
$C_3$ & $68\%,88\%$ & $95\%,99\%$ & $78\%,92\%$ & $97\%,98\%$ & $83\%,94\%$ & $97\%,99\%$\\
\hline
$C_4$ & $62\%,79\%$ & $65\%,73\%$ & $90\%,97\%$ & $79\%,85\%$ & $93\%,98\%$ & $95\%,98\%$\\
\hline
\end{tabular}
}
\vskip -5pt
\label{resulttablecifarpairtoone}
\end{table}

\section{Conclusion}
In this paper, two novel methods are proposed for targeted network fooling, namely, UPSET and ANGRI. UPSET produces a single perturbation for each class from only the target class information at the input, hence the perturbation is universal for each target class. Since only an addition and a clipping is needed to generate adversarial images, UPSET is very fast during inference. On the other hand, ANGRI has access to the input image, and hence it can produce better adversarial images as compared to UPSET for similar levels of visual fidelity. ANGRI also performs well when the classifier is trained with noisy images too, while UPSET's fooling capability degrades. Unlike \cite{baluja2017adversarial} which requires multiple networks for multiple targets, both UPSET and ANGRI are single networks that can produce targeted adversarial images but can train on multiple classification networks simultaneously, allowing them to generalize better and target multiple systems at once. Both the proposed methods are black box models, i.e. they do not require knowledge of the internals of the victim system.




{\small
\bibliographystyle{plain}
\bibliography{biblio}

\begin{thebibliography}{10}

\bibitem{baluja2017adversarial}
Shumeet Baluja and Ian Fischer.
\newblock Adversarial transformation networks: Learning to generate adversarial
  examples.
\newblock {\em arXiv preprint arXiv:1703.09387}, 2017.

\bibitem{goodfellow2014explaining}
Ian~J Goodfellow, Jonathon Shlens, and Christian Szegedy.
\newblock Explaining and harnessing adversarial examples.
\newblock {\em arXiv preprint arXiv:1412.6572}, 2014.

\bibitem{he2016deep}
Kaiming He, Xiangyu Zhang, Shaoqing Ren, and Jian Sun.
\newblock Deep residual learning for image recognition.
\newblock In {\em Proceedings of the IEEE Conference on Computer Vision and
  Pattern Recognition}, pages 770--778, 2016.

\bibitem{AdversarialExamples_KosFS17}
Jernej Kos, Ian Fischer, and Dawn Song.
\newblock Adversarial examples for generative models.
\newblock {\em CoRR}, abs/1702.06832, 2017.

\bibitem{Krizhevsky_Cifar10Citation}
Alex Krizhevsky.
\newblock {Learning Multiple Layers of Features from Tiny Images}.
\newblock Master's thesis, 2009.

\bibitem{kurakin2016adversarial}
Alexey Kurakin, Ian Goodfellow, and Samy Bengio.
\newblock Adversarial machine learning at scale.
\newblock {\em arXiv preprint arXiv:1611.01236}, 2016.

\bibitem{Kurakin2016AdversarialExamples}
Alexey Kurakin, Ian~J. Goodfellow, and Samy Bengio.
\newblock Adversarial examples in the physical world.
\newblock {\em CoRR}, abs/1607.02533, 2016.

\bibitem{MNIST_Lecun_1998}
Y.~Lecun, L.~Bottou, Y.~Bengio, and P.~Haffner.
\newblock Gradient-based learning applied to document recognition.
\newblock {\em Proceedings of the IEEE}, 86(11):2278--2324, Nov 1998.

\bibitem{TransferabelAdvExp_LiuCLS16}
Yanpei Liu, Xinyun Chen, Chang Liu, and Dawn Song.
\newblock Delving into transferable adversarial examples and black-box attacks.
\newblock {\em CoRR}, abs/1611.02770, 2016.

\bibitem{universalsegmentation}
Jan~Hendrik Metzen, Mummadi~Chaithanya Kumar, Thomas Brox, and Volker Fischer.
\newblock Universal adversarial perturbations against semantic image
  segmentation.
\newblock {\em arXiv preprint arXiv:1704.05712}, 2017.

\bibitem{moosavi2016universal}
Seyed-Mohsen Moosavi-Dezfooli, Alhussein Fawzi, Omar Fawzi, and Pascal
  Frossard.
\newblock Universal adversarial perturbations.
\newblock {\em arXiv preprint arXiv:1610.08401}, 2016.

\bibitem{DeepFool_moosavi2016}
Seyed-Mohsen Moosavi-Dezfooli, Alhussein Fawzi, and Pascal Frossard.
\newblock Deepfool: a simple and accurate method to fool deep neural networks.
\newblock In {\em Proceedings of the IEEE Conference on Computer Vision and
  Pattern Recognition}, pages 2574--2582, 2016.

\bibitem{nguyen2015deep}
Anh Nguyen, Jason Yosinski, and Jeff Clune.
\newblock Deep neural networks are easily fooled: High confidence predictions
  for unrecognizable images.
\newblock In {\em Proceedings of the IEEE Conference on Computer Vision and
  Pattern Recognition}, pages 427--436, 2015.

\bibitem{Papernot2017PracticalBlackBoxAttack}
Nicolas Papernot, Patrick McDaniel, Ian Goodfellow, Somesh Jha, Z.~Berkay
  Celik, and Ananthram Swami.
\newblock Practical black-box attacks against machine learning.
\newblock In {\em Proceedings of the 2017 ACM on Asia Conference on Computer
  and Communications Security}, ASIA CCS '17, pages 506--519, New York, NY,
  USA, 2017. ACM.

\bibitem{Papernot2016TheLO}
Nicolas Papernot, Patrick~D. McDaniel, Somesh Jha, Matt Fredrikson, Z.~Berkay
  Celik, and Ananthram Swami.
\newblock The limitations of deep learning in adversarial settings.
\newblock {\em 2016 IEEE European Symposium on Security and Privacy}, pages
  372--387, 2016.

\bibitem{Szegedy2013IntriguingPropertiesNN}
Christian Szegedy, Wojciech Zaremba, Ilya Sutskever, Joan Bruna, Dumitru Erhan,
  Ian~J. Goodfellow, and Rob Fergus.
\newblock Intriguing properties of neural networks.
\newblock {\em CoRR}, abs/1312.6199, 2013.

\bibitem{xie2017adversarial}
Cihang Xie, Jianyu Wang, Zhishuai Zhang, Yuyin Zhou, Lingxi Xie, and Alan
  Yuille.
\newblock Adversarial examples for semantic segmentation and object detection.
\newblock {\em arXiv preprint arXiv:1703.08603}, 2017.

\end{thebibliography}
}

\end{document}